\documentclass[conference]{IEEEtran}
\IEEEoverridecommandlockouts
\usepackage{cite}
\usepackage{amsmath,amssymb,amsfonts}
\usepackage{algorithmic}
\usepackage{graphicx}
\usepackage{textcomp}
\usepackage{fancyhdr}
 \usepackage{multirow}
 \usepackage{balance}
 \usepackage{makecell}
 \usepackage{hyperref}
\usepackage{xcolor}
\def\BibTeX{{\rm B\kern-.05em{\sc i\kern-.025em b}\kern-.08em
    T\kern-.1667em\lower.7ex\hbox{E}\kern-.125emX}}

\begin{document}

\title{Structured Reasoning with Tree-of-Thoughts for Bengali Math Word Problems\\
% {\footnotesize \textsuperscript{*}Note: Sub-titles are not captured in Xplore and
% should not be used}
% \thanks{Identify applicable funding agency here. If none, delete this.}
}

\author{
    \IEEEauthorblockN{
        Aurprita Mahmood \IEEEauthorrefmark{1}, 
        Sabrin alam \IEEEauthorrefmark{2}, 
        Neloy kumer Sagor \IEEEauthorrefmark{3},
        Md. Abdul Hadi \IEEEauthorrefmark{4},\\
        Md. Sehab Al Islam \IEEEauthorrefmark{5},
        Minhajul Islam \IEEEauthorrefmark{6}
    }

    \IEEEauthorblockA{
        \IEEEauthorrefmark{1}  \IEEEauthorrefmark{2}  \IEEEauthorrefmark{3}  \IEEEauthorrefmark{4} \IEEEauthorrefmark{5}  \IEEEauthorrefmark{6} Department of Computer Science and Engineering
    }
    \IEEEauthorblockA{
        \IEEEauthorrefmark{1}  \IEEEauthorrefmark{6} Ahsanullah University of Science and Technology, Dhaka, Bangladesh \\
        \IEEEauthorrefmark{2} \IEEEauthorrefmark{3} \IEEEauthorrefmark{4} \IEEEauthorrefmark{2} Southeast University, Dhaka, Bangladesh \\
    }
    \IEEEauthorblockA{
        Emails: \{\IEEEauthorrefmark{1}aurprita1234, \IEEEauthorrefmark{2}sabrinalam231,
        \IEEEauthorrefmark{3}neloysagor12134,
        \IEEEauthorrefmark{4}mdhadi8678,
        \IEEEauthorrefmark{5}sehabalislam,
        \IEEEauthorrefmark{6}minhajulislam.aust\}@gmail.com
    }
    % \IEEEauthorblockA{
    %     \IEEEauthorrefmark{1} \textit{Corresponding Author}
    % }
}

\maketitle

\begin{abstract}Mathematical Word Problems (MWPs) are among the most challenging tasks in natural language processing because they require both linguistic understanding and multi-step numerical reasoning. While Chain-of-Thought (CoT) prompting has shown promise, its linear structure often propagates errors, limiting overall effectiveness. To address this limitation, we present the a systematic study of Tree-of-Thought (ToT) reasoning for Bengali MWPs using the SOMADHAN dataset. Owing to computational and token-cost constraints, we evaluate a curated set of 100 representative problems across multiple large language models (LLMs), including GPT-OSS and LLaMA variants, under standard prompting, CoT, and ToT strategies. Our results show that CoT improves baseline accuracy from 78\% (standard prompting) to 83\% on average, while ToT further increases performance by up to 5 percentage points, achieving 88\% accuracy with GPT-OSS-120B. These improvements highlight that ToT is particularly effective in medium-to-large-scale models but may offer less advantage for smaller ones. Overall, our findings establish ToT as a robust framework for solving mathematical problems in low-resource languages such as Bengali. More broadly, this study shows that structured reasoning methods like ToT can provide more reliable and globally consistent outcomes than CoT, paving the way for better reasoning strategies in multilingual NLP.

%Mathematical Word Problems (MWPs) are among the most challenging tasks in natural language processing because they require both linguistic understanding and multi-step numerical reasoning. While Chain-of-Thought (CoT) prompting has shown promise, its linear structure often propagates errors, limiting effectiveness. To address this, we present the first systematic study of Tree-of-Thought (ToT) reasoning for Bengali MWPs using the SOMADHAN dataset. Due to resource constraints, we evaluated 100 representative problems across multiple large language models (LLMs), including GPT-OSS and LLaMA variants, under standard prompting, CoT, and ToT strategies. Our results show that CoT improves baseline accuracy from 78\% (standard prompting) to 83\% on average, while ToT further increases performance by up to 5 percentage points, achieving 88\% accuracy with GPT-OSS-120B. These findings demonstrate that ToT provides more reliable and globally consistent reasoning compared to CoT, especially in medium-to-large-scale models, and establish ToT as a robust framework for solving mathematical problems in low-resource languages such as Bengali.
\end{abstract}


\section{Introduction}

Mathematical Word Problems (MWPs) are among the most demanding tasks in natural language processing (NLP), as they require simultaneous comprehension of natural language, identification of numerical relationships, and multi-step logical reasoning. In recent years, large language models (LLMs) have demonstrated remarkable improvements in solving such problems through advanced prompting techniques. Among these, Chain-of-Thought (CoT) prompting has proven effective by guiding models to generate intermediate reasoning steps before producing a final answer. However, CoT follows a single linear reasoning trajectory, and an early error within this chain often propagates throughout the solution process, resulting in an incorrect final outcome.

This limitation has motivated the development of more structured reasoning paradigms. One such approach is Tree-of-Thought (ToT) reasoning, which enables a model to explore multiple candidate reasoning paths in parallel, evaluate intermediate steps, and dynamically revise or select the most consistent solution. By introducing branching and backtracking mechanisms, ToT provides a more flexible and globally consistent reasoning framework than conventional linear approaches.

Although structured reasoning methods such as CoT and ToT have demonstrated success on English-language benchmarks, including GSM8K and MATH, research focusing on low-resource languages, particularly Bengali, remains limited. The recent release of the SOMADHAN dataset, which contains 8,792 complex Bengali MWPs accompanied by step-by-step solutions, has created a valuable opportunity to investigate advanced reasoning strategies in this linguistic context. However, existing studies on Bengali MWPs have primarily relied on CoT or direct prompting, and to the best of our knowledge, Tree-of-Thought reasoning has not yet been systematically evaluated for Bengali mathematical problems.

In this paper, we present the first comprehensive exploration of Tree-of-Thought reasoning for Bengali MWPs. Using a representative subset of the SOMADHAN dataset, we evaluate several large language models under standard prompting, Chain-of-Thought, and Tree-of-Thought strategies. The primary objectives of this study are to investigate:  
(i) the extent to which ToT improves performance over CoT for Bengali MWPs,  
(ii) the influence of model scale on the effectiveness of ToT, and  
(iii) the types of problems that benefit most from structured, tree-based reasoning.

Through this analysis, we aim to establish a benchmark for structured reasoning in the Bengali language and highlight the broader potential of Tree-of-Thought frameworks for low-resource and multilingual NLP applications.

\begin{figure*}[h]
\centering
\includegraphics[width=0.8\textwidth]{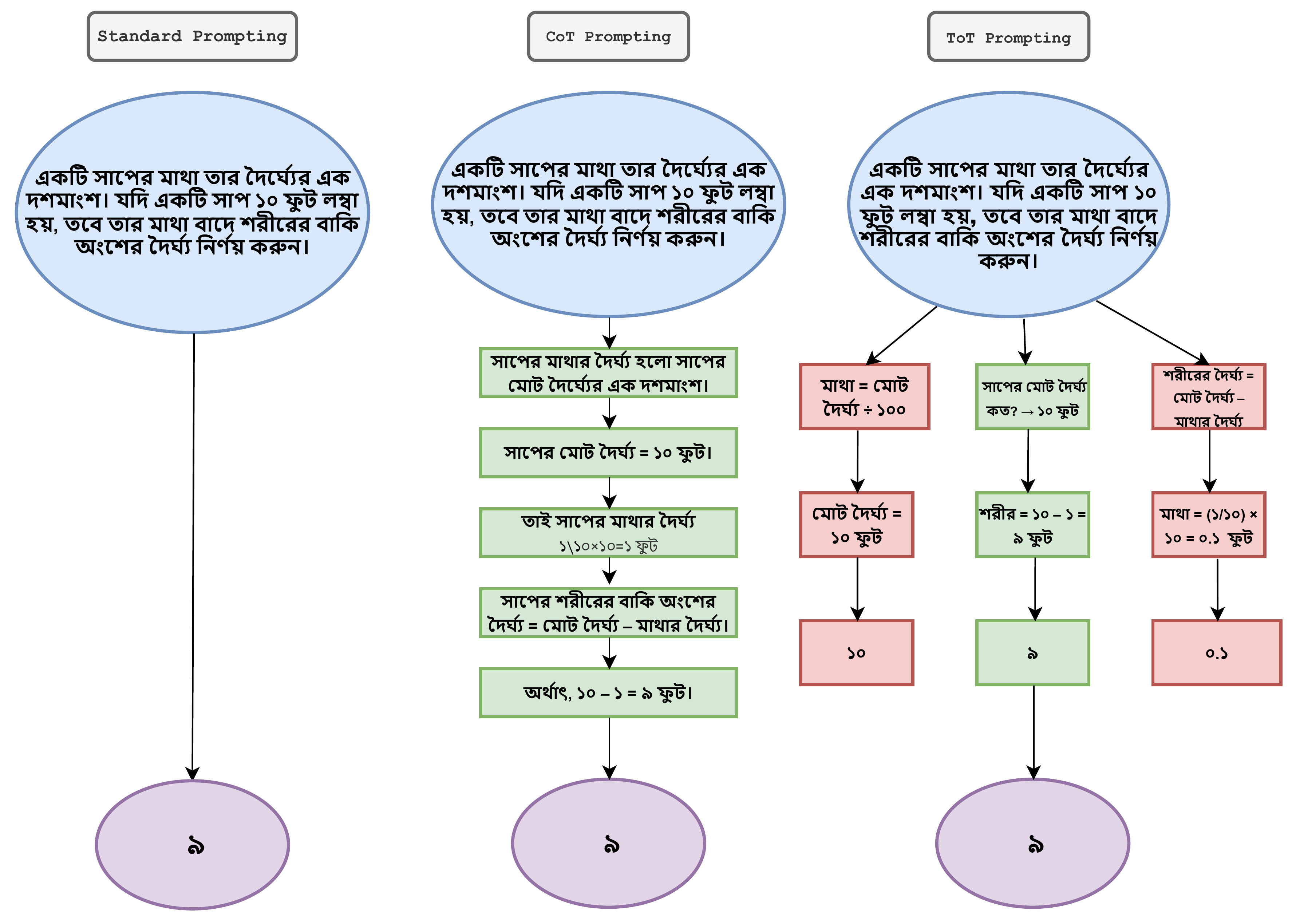} 
\caption{Comparison of Standard Prompting, Chain-of-Thought (CoT) Prompting, and Tree-of-Thought (ToT) Prompting in Bengali Mathematical Reasoning}
\label{fig:intro}
\end{figure*}

Motivated by these findings, this study presents a systematic study of Tree-of-Thought reasoning for Bengali MWPs. Using the SOMADHAN dataset as a benchmark, we evaluate state-of-the-art LLMs, including GPT-4o, GPT-3.5, LLaMA-3, and Llama-4 under both CoT and ToT prompting strategies. Specifically, we investigate three key questions: (i) to what extent does ToT improve reasoning accuracy over CoT in Bengali MWPs; (ii) how do different LLMs vary in their ability to exploit ToT reasoning; and (iii) what types of MWPs benefit most from tree-structured reasoning. (iv) how does the number of few-shot examples influence the effectiveness of CoT in Bengali MWPs. Through this comparative analysis, we aim to establish a new benchmark for reasoning in low-resource languages and demonstrate the value of ToT beyond English-centric datasets. The main contributions of this paper are as follows:

\begin{enumerate}
    \item We introduce the ToT-based evaluation framework for Bengali mathematical word problems.
    \item We applied various few-shot prompting strategies to further analyze CoT, aiming to determine the optimal number of examples required for better accuracy.
    \item We provide a comparative analysis of CoT and ToT prompting across several LLMs, showing that ToT achieves superior accuracy, particularly for complex multi-step MWPs.
    \item We release benchmark results and analysis to foster further research on reasoning in low-resource languages.
\end{enumerate}

\section{\textbf{Related Work}}

Our review encompasses a diverse set of contributions from the domains of language model reasoning, and Bangla natural language processing. In what follows, we analyze each study in terms of its core approach, dataset utilization, and empirical performance, thereby identifying the research gaps that motivate our own work.

Liang et al.~\cite{liang2025cotmath} presented a theoretical foundation for Chain-of-Thought (CoT) reasoning by modeling it as a Markov process and analyzing it through information theory. Using datasets such as e-SNLI, ANLI, CommonsenseQA, and SVAMP, they demonstrated that CoT prompting significantly boosts reasoning accuracy. For instance, few-shot arithmetic reasoning accuracy improved from 4\% to 70\% with one example and up to 90\% with four, far surpassing direct prompting.

% Vali~\cite{vali2013new} introduced a hybrid framework combining the Subdivision Labeling Method (SLM) with a Clustering-Based Parallel Genetic Algorithm (CBPGA). Tested on benchmark functions such as Rosenbrock, Shekel’s Foxholes, and Easom, the method improved convergence speed and accuracy, outperforming Random Search, Simulated Annealing, and traditional GA models with efficiency gains above 50\%.

% Mukherjee et al.~\cite{mitra2023orca} proposed Orca 2, a reasoning-centric model that extends small language model capabilities beyond simple imitation. The approach integrates multiple reasoning strategies such as step-by-step reasoning, recall-then-generate, and cautious reasoning, alongside a novel technique called Prompt Erasing. The system was evaluated on about 36K prompts across 15 benchmarks. Reported accuracies included 91.0\% on GSM8K, 86.3\% on SVAMP, 83.5\% on ASDiv, 54.7\% on AQuA, and 56.0\% on the MATH dataset, outperforming larger models through strategic reasoning instruction. On the other side, Zhang et al.~\cite{guan2025rstar} developed rStar-Math, a compact language model (~1.3B parameters) designed for mathematical reasoning. The approach applies a three-phase self-evolution strategy, including imitation learning, recursive self-generated reasoning, and hallucination detection for error correction. The framework was evaluated on GSM8K, SVAMP, MultiArith, and MATH datasets. Reported results demonstrated 91.6\% accuracy on GSM8K, 89.2\% on SVAMP, 99.4\% on MultiArith, and 50.4\% on MATH, establishing new state-of-the-art results among small LLMs.

Xu et al.~\cite{xu2025colength} investigated the impact of context length on math word problem solving by introducing the Extended Grade-School Math (E-GSM) dataset. They evaluated four proprietary and four open-source LLMs under CoT prompting and found that performance degraded significantly on longer MWPs. To address this, they proposed Condition-Retrieving Instruction (CoRe) for proprietary models and auxiliary-task fine-tuning for open-source models, achieving notable accuracy gains across E-GSM, GSM-IC, and SVAMP. I another study, Long~\cite{long2023tot} proposed the Large Language Model Guided Tree-of-Thought (ToT) framework, enhancing reasoning by augmenting LLMs with a prompter, checker, memory, and ToT controller. The system was applied to Sudoku puzzle solving, where it significantly outperformed standard CoT by enabling backtracking and multi-path exploration. Results demonstrated a higher success rate in complex reasoning tasks compared to linear prompting.

% Bryant et al.~\cite{bryant2008mathintervention} studied early math interventions using a regression discontinuity design with 266 first- and second-grade students identified with math difficulties. Tier 2 students received 15-minute booster lessons for 18 weeks. Results showed significant improvements for second graders on the TEMI-PM benchmark, while gains for first graders were not statistically significant. 

In the domain of Graph-of-Thoughts, Bai et al.~\cite{bai2025sagot} introduced SaGoT (Self-attention-based Graph-of-Thought), a decoding method that enables LLMs to generate reasoning steps in a graph structure instead of linear chains. Unlike prompting-based graph methods, SaGoT modifies the transformer’s self-attention to construct a reasoning graph during inference. Experiments showed consistent accuracy improvements in solving complex math problems while enhancing interpretability.

Zhu et al.~\cite{zhu2025chain} proposed LONGREPS, a process-supervised framework designed to enhance long-context reasoning in LLMs through Chain-of-Thought (CoT) path supervision. Using the MuSiQue dataset as the primary benchmark, the method combines self-sampling of reasoning paths with a novel quality assessment protocol emphasizing answer correctness, source faithfulness, and intrinsic consistency. Experimental results demonstrated substantial gains, with LLaMA-3.1-8B improving by +13.6 points and Qwen-2.5-7B by +3.8 points on MuSiQue, and average improvements of +9.3/+8.1 across diverse QA tasks in LongBenchV1 and V2, confirming the effectiveness of reasoning path supervision for long-context tasks .

Taurpa et al.~\cite{aurpa2024shomikoron} presented Shomikoron, a dataset of 3,430 Bangla mathematical statements paired with equations. The goal was to support equation extraction in Bangla, a low-resource language. Transformer-based models including BanglaBERT and XLM-RoBERTa were applied, achieving an accuracy of 98.95\%, thereby establishing Shomikoron as a high-quality benchmark for Bangla mathematical text processing. Likewise, Era at al.~\cite{era2024empowering}, presented PatiGonit dataset, establishing a foundation for systematic research on Bengali mathematical text by introducing a standardized benchmark for model evaluation. It facilitated exploration of both transformer-based architectures and neural machine translation (NMT) approaches for converting natural language problem statements into formal mathematical equations.

Noguer i Alonso~\cite{alonso2024math} developed a unified theoretical framework for Chain-of-Thought (CoT) and Tree-of-Thought (ToT) reasoning in transformers. Using tools from computational complexity, measure theory, and numerical analysis, the study provided formal proofs of CoT’s ability to expand transformer expressiveness and ToT’s capacity for branching search. Empirical validation on the ARC dataset confirmed up to 6× improvements in reasoning with optimized test-time strategies. Sprague et al.~\cite{sprague2025tocot} conducted a meta-analysis of over 100 CoT studies and additional experiments across 20 datasets with 14 LLMs. Their findings showed that CoT provides substantial gains primarily in math and symbolic reasoning tasks, with improvements of 12.3\% on math and 14.2\% on symbolic datasets, but minimal benefits on commonsense or encyclopedic tasks. This suggests CoT should be applied selectively to domains requiring formal computation.

Despite notable progress in mathematical NLP, research on Bangla math word problems remains limited. Existing studies primarily rely on Chain-of-Thought (CoT) prompting, which improves reasoning but has clear limitations. To the best of our knowledge, Tree-of-Thought (ToT) reasoning has not yet been applied in this domain, presenting a valuable opportunity for developing more structured and effective problem-solving approaches.

\section{\textbf{Methodology}}
\begin{figure}[h]
\centering
\includegraphics[width=0.52\columnwidth]{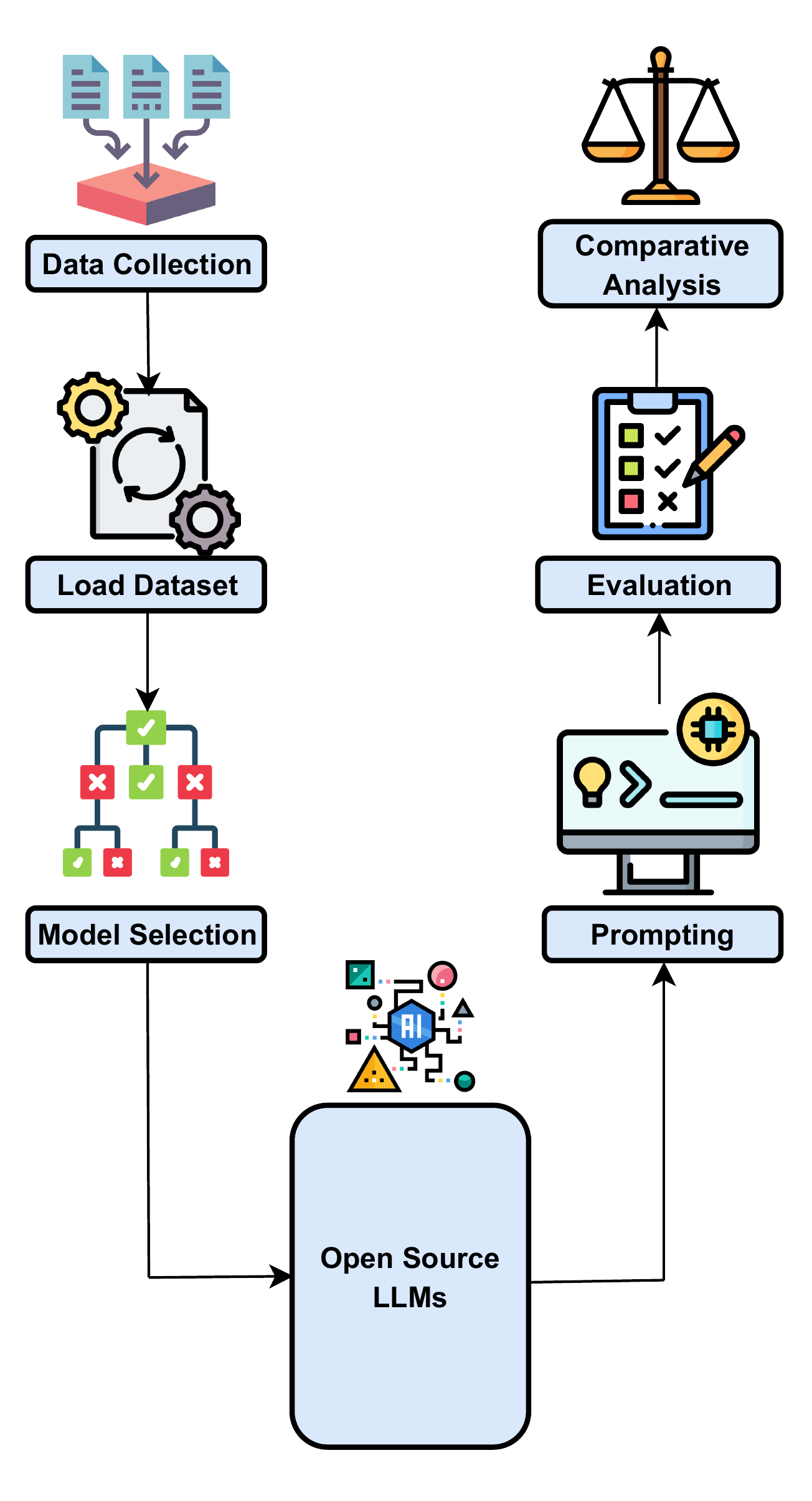} 
\caption{Proposed Methodology}
\label{fig:meth}
\end{figure}

We evaluate Bengali math word problems using the SOMADHAN dataset and a range of large language models. An illustration of our proposed methodology is given in Figure~\ref{fig:meth}.

\subsection{Dataset}

\begin{figure}[h]
\centering
\includegraphics[width=1.0\columnwidth]{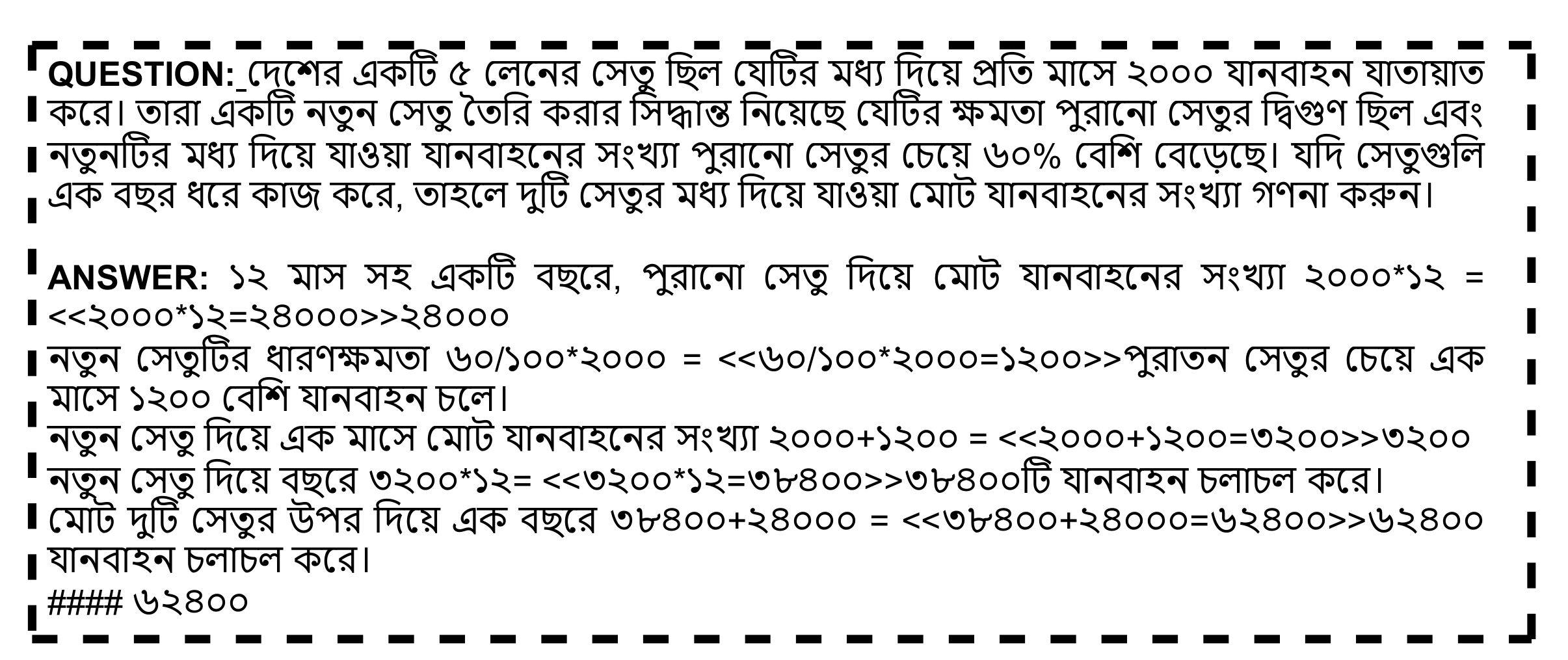} 
\caption{A sample of SOMADHAN Dataset}
\label{fig:data}
\end{figure}

For this work, we employ the \href{https://data.mendeley.com/datasets/34bs5cxk9j/2}{SOMADHAN}
 dataset~\cite{paul2025leveraging}, which contains 8,792 Bengali complex mathematical word problems with solutions in CSV format. Each entry consists of two fields: a question written in Bengali and a corresponding answer that provides a solved solution. The answers are structured to include both intermediate reasoning steps and the final result, where intermediate calculations are marked and the final solution is explicitly indicated with the prefix. Unlike earlier Bengali mathematical resources that focus primarily on entity recognition or equation extraction, this dataset emphasizes complex word problems with step-by-step reasoning. As such, it provides a valuable benchmark for evaluating reasoning-oriented models and is particularly suited to exploring advanced strategies like Tree-of-Thought (ToT), which has not yet been applied in this domain. An example of the SOMADHAN dataset has been shown in Figure~\ref{fig:data}.

\subsection{Models}
In this study, we evaluate both open-source large language models (LLMs) under Chain-of-Thought (CoT) and Tree-of-Thought (ToT) prompting strategies along with standard prompting. This dual perspective allows us to compare reasoning capabilities across models of varying parameter sizes, training paradigms, and accessibility. We categorize the models into two groups: OpenAI models and LLaMA models.

\subsubsection{OpenAI Models}
We evaluate two recently released open-weight reasoning models from the GPT-OSS family, both optimized for symbolic and mathematical problem solving through alignment and extended context training~\cite{openai2025gptoss}. The first, GPT-OSS-20B, is a mid-sized 20B parameter model that balances computational efficiency with robust reasoning accuracy. Prior benchmarks demonstrate that GPT-OSS-20B achieves competitive performance on datasets such as GSM8K and MATH when paired with Chain-of-Thought prompting~\cite{wei2022chain}, making it an ideal baseline for examining the limitations of linear reasoning. The second, GPT-OSS-120B, is a large-scale 120B parameter model that represents the upper bound of OpenAI’s reasoning-oriented architectures. Its larger capacity enables stronger performance on long-context reasoning and complex multi-step tasks. We selected these models because they provide complementary perspectives: CoT allows us to establish baseline reasoning accuracy, while ToT enables branching exploration that tests whether larger models can leverage structured search to mitigate error propagation and achieve globally consistent solutions~\cite{yao2023tree,haji2024multiagenttot}. Together, they allow us to systematically analyze how parameter scaling interacts with structured prompting strategies in solving Bengali mathematical word problems.

\subsubsection{LLaMA Models}
We further evaluate open-source LLaMA models, which provide transparent baselines for reasoning and multilingual tasks~\cite{touvron2023llama,meta2024llama3}. Four variants are considered, representing different parameter sizes and design philosophies. The lightweight LLaMA-3.1-8B-Instant enables us to examine how CoT establishes baseline reasoning under constrained capacity and whether ToT can improve performance in small-scale models. At the other end, the LLaMA-3.3-70B-Versatile achieves state-of-the-art results on multilingual and mathematical reasoning benchmarks~\cite{meta2024llama3}, serving as our open-source upper bound to compare the effectiveness of ToT versus CoT at scale. We also include two instruction-tuned LLaMA-4 variants: Maverick-17B-128E, which leverages expanded embeddings for improved reasoning stability, and Scout-17B-16E, which prioritizes efficiency with reduced embeddings. Both are tested under CoT for alignment in step-by-step problem solving and ToT for branching robustness, allowing us to study trade-offs between reasoning accuracy and computational cost across diverse open-source architectures.

\begin{table*}[t]
\centering
\caption{Accuracy of LLMs (\%) on Bengali MWPs under Standard Prompting, CoT (1/2/5/7-shot), and ToT.}
\label{tab:cot_tot}
{ % start local group for arraystretch
\renewcommand{\arraystretch}{1.6}
\begin{tabular}{|l|c|cc|cc|cc|cc|cc|}
\hline
\multirow{2}{*}{\textbf{Model}} & \multirow{2}{*}{\makecell{\textbf{Standard} \\ \textbf{Prompting}\\ \textbf{Zero Shot}}} 
& \multicolumn{8}{c|}{\textbf{Chain of Thought (CoT)}} 
& \multicolumn{2}{c|}{\textbf{Tree of Thought}} \\
\cline{3-10}
 &  & \multicolumn{2}{c|}{\textbf{1-shot}} 
    & \multicolumn{2}{c|}{\textbf{2-shot}} 
    & \multicolumn{2}{c|}{\textbf{5-shot}} 
    & \multicolumn{2}{c|}{\textbf{7-shot}} 
    & \multicolumn{2}{c|}{\textbf{Zero Shot}}  \\
\cline{3-10}
 &  & \textbf{Med.} & \textbf{High} 
    & \textbf{Med.} & \textbf{High} 
    & \textbf{Med.} & \textbf{High} 
    & \textbf{Med.} & \textbf{High} 
    & \textbf{Med.} & \textbf{High}  \\
\hline
GPT-OSS-20B                 & 78.00 & 82.00 & 83.00 & 84.00 & 84.00 & 86.00 & 88.00 & 82.00 & 84.00 & \textbf{87.00} & 84.00 \\
GPT-OSS-120B                & 80.00 & 84.00 & 85.00 & 85.00 & 86.00 & 85.00 & 87.00 & 82.00 & 84.00 & \textbf{88.00} & 86.00 \\
\hline
LLaMA-3.1-8B-instant         & 48.00 & \multicolumn{2}{c|}{51.00} & \multicolumn{2}{c|}{51.00} & \multicolumn{2}{c|}{41.00} & \multicolumn{2}{c|}{\textbf{73.00}} & \multicolumn{2}{c|}{31.00} \\
LLaMA-3.3-70B-versatile      & 79.00 & \multicolumn{2}{c|}{85.00} & \multicolumn{2}{c|}{86.00} & \multicolumn{2}{c|}{86.00} & \multicolumn{2}{c|}{87.00} & \multicolumn{2}{c|}{\textbf{88.00}} \\
LLaMA-4-maverick-17B-128E    & 84.00 & \multicolumn{2}{c|}{84.00} & \multicolumn{2}{c|}{85.00} & \multicolumn{2}{c|}{83.00} & \multicolumn{2}{c|}{83.00} & \multicolumn{2}{c|}{\textbf{88.00}} \\
LLaMA-4-scout-17B-16E        & 79.00 & \multicolumn{2}{c|}{76.00} & \multicolumn{2}{c|}{86.00} & \multicolumn{2}{c|}{82.00} & \multicolumn{2}{c|}{82.00} & \multicolumn{2}{c|}{\textbf{87.00}} \\
\hline
\end{tabular}}
\end{table*}

We employ both proprietary GPT-OSS models and open-source LLaMA models to ensure a balanced evaluation of Chain-of-Thought (CoT) and Tree-of-Thought (ToT) prompting. The GPT-OSS series provides reasoning-optimized baselines with strong arithmetic performance, while the LLaMA family offers transparent, diverse parameter scales ranging from lightweight to large instruction-tuned variants. By testing these models under both CoT and ToT, we aim to (i) measure how structured prompting impacts reasoning across different model sizes, and (ii) establish whether open-source LLaMA models can match or exceed proprietary GPT models in solving Bengali mathematical word problems.

\section{\textbf{Prompting Strategies}}
In this study, we evaluated different prompting strategies to compare the effectiveness of standard prompting, chain-of-thought (CoT) prompting, and our proposed tree-of-thoughts (ToT) prompting. The strategies were designed to test both zero-shot and few-shot settings, allowing us to examine the impact of additional exemplars on model reasoning performance.

\subsection{Zero-Shot Prompting}
\begin{figure}[h]
\centering
\includegraphics[width=1.0\columnwidth]{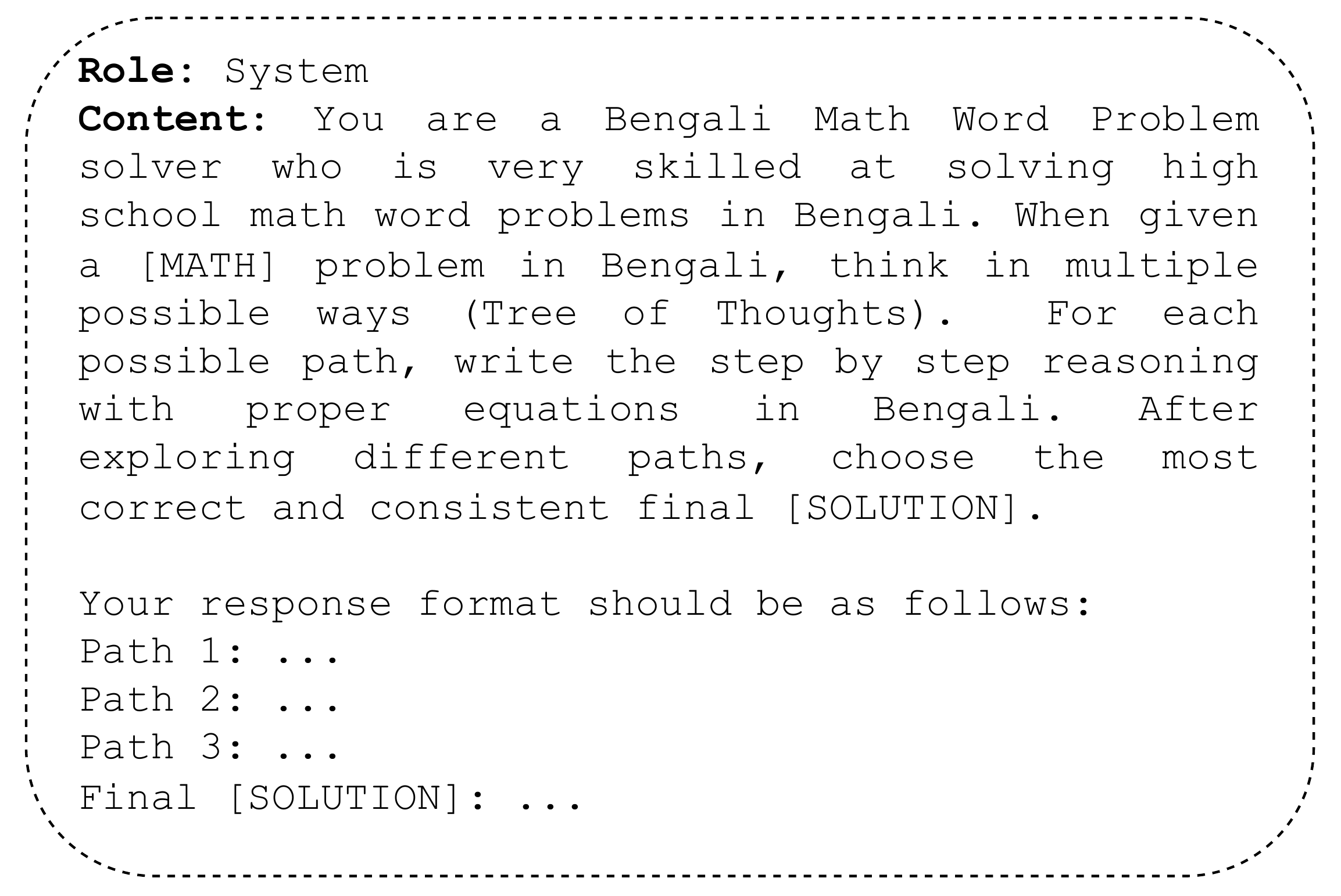} 
\caption{System instruction for the tree-of-thoughts (ToT) prompting strategy.}
\label{fig:system}
\end{figure}

For standard prompting, we used only the zero-shot setting, where the model was directly asked to solve the problem without being given any examples. The prompting structure for this setup was adapted from Paul et al.~\cite{paul2025leveraging}. Similarly, our proposed tree-of-thoughts prompting was evaluated exclusively under the zero-shot setting. In this case, the model explored multiple reasoning branches before converging on the final solution, and the system instruction for it is shown in Figure~\ref{fig:system}.

\subsection{Few-Shot Prompting}
For chain-of-thought prompting, we employed multiple few-shot settings to assess how providing different numbers of exemplars influences reasoning performance. Specifically, we tested the 1-shot, 2-shot, 5-shot, and 7-shot setups, where worked-out examples were included in the prompt to guide the model’s step-by-step reasoning. The prompting structures for these settings were also adapted from Paul et al. (2025)~\cite{paul2025leveraging}.

\subsection{Experimental Setup}
We conducted all experiments on Google Colab, a cloud-based platform providing up to 12 GB RAM, a 2-core CPU, and GPU acceleration (Tesla T4/V100 with 16 GB VRAM). Model inference was carried out through the Groq API, which provided access to both GPT-OSS and LLaMA models. Instead of using the full SOMADHAN dataset, we evaluated a subset of 100 math word problems for our closed-source LLMs due to token limit constraints.

%We conducted all experiments on Google Colab, a cloud-based platform providing up to 12 GB RAM, a 2-core CPU, and GPU acceleration (Tesla T4/V100 with 16 GB VRAM). Model inference was carried out through the Groq API, which provided access to both GPT-OSS and LLaMA models. A subset of 100 math word problems was selected from the SOMADHAN dataset for evaluation to ensure feasibility under manual assessment. This restriction was necessary due to resource and token constraints, which made full-dataset evaluation impractical within our setup. Since the task involved multi-step reasoning, all outputs were evaluated by human annotators based on final-answer correctness rather than automated scripts. This setup ensured a reproducible, cost-effective, and human-verified benchmark for comparing CoT and ToT prompting across different model scales.

%We conducted all experiments on Google Colab, a cloud-based platform providing up to 12 GB RAM, a 2-core CPU, and GPU acceleration (Tesla T4/V100 with 16 GB VRAM). Model inference was carried out through the Groq API, which provided access to both GPT-OSS and LLaMA models. A subset of 100 math word problems was selected from the SOMADHAN dataset for evaluation to ensure feasibility under manual assessment. Since the task involved multi-step reasoning, all outputs were evaluated by human annotators based on final-answer correctness rather than automated scripts. This setup ensured a reproducible, cost-effective, and human-verified benchmark for comparing CoT and ToT prompting across different model scales.

\subsection{Hyperparameter Settings}
For the GPT-OSS models, we fixed the temperature at 1.0, which proved both convenient and reliable. We also varied the reasoning capability between high and medium to assess differences in reasoning depth. For the LLaMA models, the temperature was likewise set to 1.0, ensuring consistency across all evaluations. The maximum token limit was fixed at 1024 for all models. This uniform setup allowed performance differences to be attributed mainly to the prompting strategy (CoT vs. ToT) and model architecture.

\subsection{Evaluation Metric}
Since multi-step reasoning cannot always be evaluated step by step, we assessed model performance based solely on the final answer correctness. A prediction was considered correct if the final numeric or symbolic result exactly matched the ground truth. All outputs were manually validated by human evaluators, and no automated scoring scripts were used. This evaluation protocol aligns with prior studies on Bengali MWPs, where Paul et al.~\cite{paul2025leveraging} emphasized final-answer validation as the principal benchmark for assessing large language models.

\begin{figure}[h]
\centering
\includegraphics[width=1.0\columnwidth]{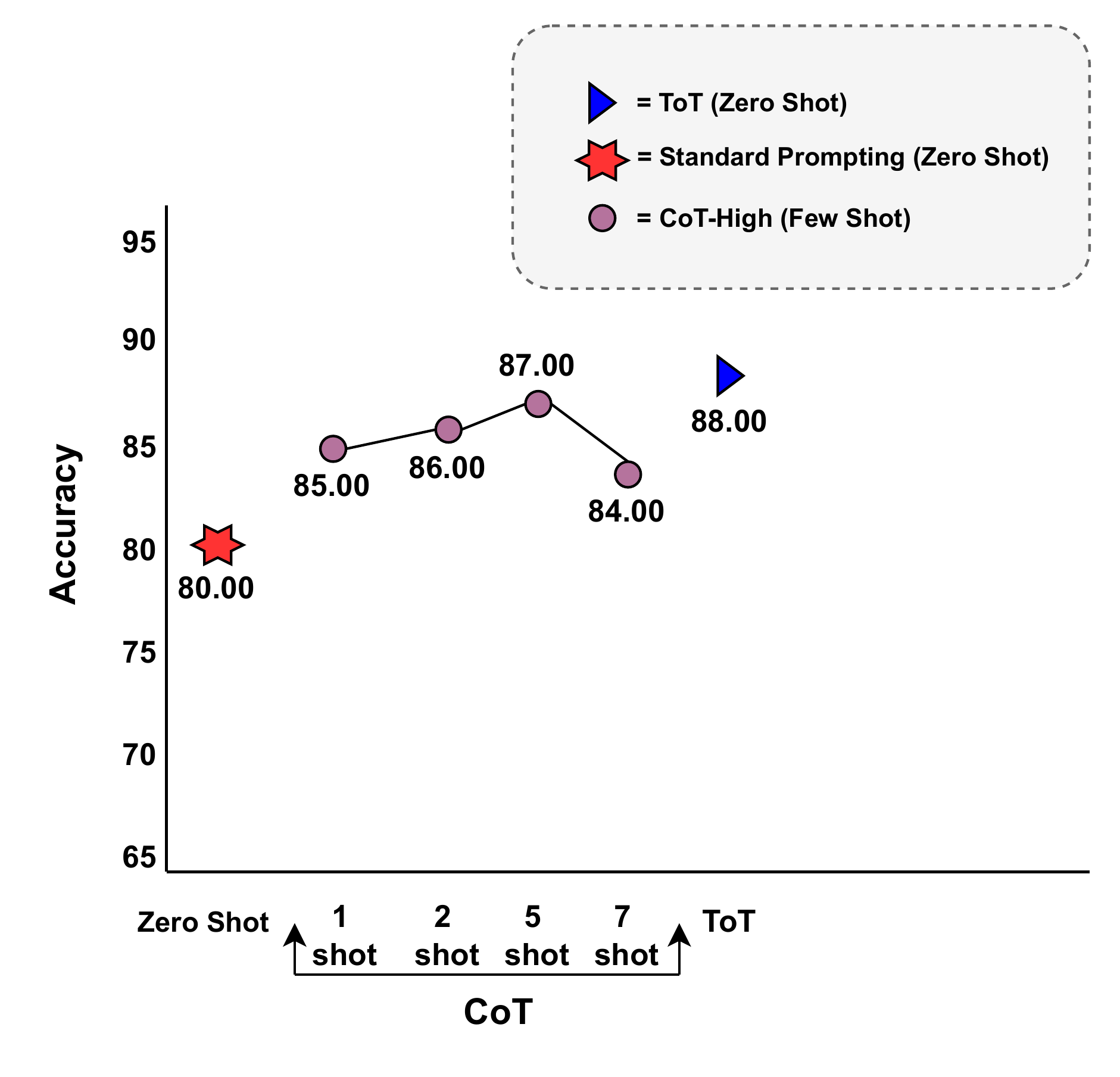} 
\caption{Accuracy of Best Performing Model (GPT-OSS-120B) on three different approaches}
\label{fig:graph}
\end{figure}

\section{\textbf{Result Analysis}}
The results in Table~\ref{tab:cot_tot} highlight several notable trends across prompting strategies and model families. Standard prompting consistently produced the lowest accuracies, with performance ranging from 48\% for LLaMA-3.1-8B to 84\% for LLaMA-4-Maverick. This confirms that unguided prompting is inadequate for solving Bengali mathematical word problems, as it fails to capture the step-by-step reasoning required for correct solutions.

Chain-of-Thought (CoT) prompting led to significant improvements across most models. Accuracy generally increased with the number of shots, though gains diminished beyond the 5-shot setting. For example, GPT-OSS-20B improved from 78\% under standard prompting to 86\% in the 5-shot CoT setting, while GPT-OSS-120B peaked at 88\% in the 1-shot high configuration. Among the open-source models, LLaMA-3.3-70B demonstrated stable performance, maintaining between 85\% and 87\% across CoT variations. LLaMA-4-Maverick also achieved strong results, peaking at 86\% under 5-shot CoT. In contrast, LLaMA-3.1-8B showed highly unstable behavior, improving to 73\% at 7-shot CoT but dropping to as low as 41\% at 5-shot, suggesting that smaller models struggle to sustain consistent reasoning chains.

Tree-of-Thought (ToT) prompting further enhanced performance in most medium and large models, confirming the advantages of structured exploration. GPT-OSS-20B improved from its CoT peak of 86\% to 87\% with ToT, while GPT-OSS-120B maintained its best performance at 88\% under ToT. Similarly, LLaMA-3.3-70B increased from 87\% with CoT to 88\% with ToT. Both LLaMA-4 variants also benefited: Maverick improved from 83\% in CoT to 88\% under ToT, and Scout rose from 82\% to 87\%. These results indicate that larger, instruction-tuned models are able to fully exploit the branching reasoning offered by ToT. In sharp contrast, LLaMA-3.1-8B collapsed under ToT, falling to just 31\%, which reinforces that smaller models lack the representational capacity to benefit from tree-based reasoning.
Overall, the findings in Table~\ref{tab:cot_tot} highlight a clear progression across prompting strategies: standard prompting is insufficient, CoT provides notable improvements but remains unstable in smaller models, and ToT consistently achieves superior robustness and accuracy in larger architectures. In particular, ToT outperformed CoT by up to five percentage points, with GPT-OSS-120B reaching its peak accuracy of 88\% under ToT, surpassing both zero-shot standard prompting and all CoT settings (Fig.~\ref{fig:graph}). These results confirm ToT as the most reliable prompting strategy for enabling globally consistent reasoning in Bengali mathematical word problems.

\section{\textbf{Limitations}}
This study was limited to 100 samples from the SOMADHAN dataset, which may not fully capture its linguistic and numerical diversity. We also relied solely on open-access models, potentially underestimating the performance of stronger closed-source systems. Moreover, smaller models showed unstable behavior, indicating that the effectiveness of Tree-of-Thought prompting depends heavily on model scale and architecture. Future work should expand evaluation to the full dataset and a broader range of model families.

%The experiments in this study were conducted on a limited subset of 100 samples, which may not fully capture the linguistic, numerical, and contextual diversity of the SOMADHAN dataset. As a result, the reported accuracies may not generalize across all problem types or difficulty levels. Furthermore, we relied exclusively on open-access models due to resource constraints, while recent studies have shown that closed-source commercial systems often achieve stronger and more consistent reasoning performance. This reliance on open models restricts the scope of comparison and may underestimate the state-of-the-art. In addition, smaller models demonstrated highly unstable behavior, highlighting that the effectiveness of Tree-of-Thought prompting is strongly dependent on model scale and architecture. Future work should therefore expand evaluation to larger datasets, include a broader range of model famil

\section{\textbf{Conclusion and Future Work}}

This work provides the first systematic evaluation of Tree-of-Thought (ToT) reasoning for Bengali mathematical word problems. Our findings show that ToT not only surpasses standard prompting but also consistently matches or improves upon Chain-of-Thought (CoT), particularly in larger models, by enabling branching exploration and reducing error propagation. While the study demonstrates clear advantages of ToT, the evaluation was limited to 100 samples and open-access models due to resource constraints, which may not fully capture the diversity of the SOMADHAN dataset or the capabilities of closed-source systems. Future research should therefore extend experiments to the complete dataset, incorporate state-of-the-art proprietary models, and investigate hybrid reasoning frameworks such as validator-augmented ToT, Graph-of-Thought, or multi-agent systems. Such directions hold promise for advancing mathematical reasoning in low-resource languages and supporting applications in education and intelligent tutoring.

\balance

\bibliographystyle{IEEEtran} % or plain, unsrt, etc.
\bibliography{references}

@article{paul2025leveraging,
  title={Leveraging Large Language Models for Bengali Math Word Problem Solving with Chain of Thought Reasoning},
  author={Paul, Bidyarthi and Era, Jalisha Jashim and Zim, Mirazur Rahman and Aothoi, Tahmid Sattar and Shah, Faisal Muhammad},
  journal={arXiv preprint arXiv:2505.21354},
  year={2025}
}

@article{wei2022chain,
  title={Chain-of-thought prompting elicits reasoning in large language models},
  author={Wei, Jason and Wang, Xuezhi and Schuurmans, Dale and Bosma, Maarten and Xia, Fei and Chi, Ed and Le, Quoc V and Zhou, Denny and others},
  journal={Advances in neural information processing systems},
  volume={35},
  pages={24824--24837},
  year={2022}
}

@article{yao2023tree,
  title={Tree of thoughts: Deliberate problem solving with large language models},
  author={Yao, Shunyu and Yu, Dian and Zhao, Jeffrey and Shafran, Izhak and Griffiths, Tom and Cao, Yuan and Narasimhan, Karthik},
  journal={Advances in neural information processing systems},
  volume={36},
  pages={11809--11822},
  year={2023}
}

@article{aurpa2024shomikoron,
  title={Shomikoron: Dataset to discover equations from Bangla Mathematical text},
  author={Aurpa, Tanjim Taharat and Fariha, Kazi Noshin and Hossain, Kawser},
  journal={Data in Brief},
  volume={55},
  pages={110742},
  year={2024},
  publisher={Elsevier}
}

@article{zhu2025chain,
  title={Chain-of-thought matters: improving long-context language models with reasoning path supervision},
  author={Zhu, Dawei and Wei, Xiyu and Zhao, Guangxiang and Wu, Wenhao and Zou, Haosheng and Ran, Junfeng and Wang, Xun and Sun, Lin and Zhang, Xiangzheng and Li, Sujian},
  journal={arXiv preprint arXiv:2502.20790},
  year={2025}
}

@inproceedings{liang2025cotmath,
  title={Chain-of-Thought Reasoning for Math: Theoretical Foundation and Applications},
  author={Liang, Jessica E.},
  booktitle={Proceedings of the 2nd AI for MATH Workshop, ICML},
  year={2025},
  address={Vancouver, Canada}
}

@inproceedings{xu2025colength,
  title={Can LLMs Solve Longer Math Word Problems Better?},
  author={Xu, Xin and Xiao, Tong and Chao, Zitong and Huang, Zhenya and Yang, Can and Wang, Yang},
  booktitle={Proceedings of ICLR},
  year={2025},

}

@misc{long2023tot,
  title={Large Language Model Guided Tree-of-Thought},
  author={Long, Jieyi},
  year={2023},
  howpublished={arXiv preprint arXiv:2305.08291},
 
}

@inproceedings{bai2025sagot,
  title={Self-attention-based Graph-of-Thought for Math Problem Solving},
  author={Bai, Ruiqiao and Han, Xue and Lei, Shuo and Feng, Junlan and Luo, Yanyan and Deng, Chao},
  booktitle={Findings of the Association for Computational Linguistics: ACL},
  pages={6112--6125},
  year={2025}
}

@misc{alonso2024math,
  title={The Mathematics of Chain of Thought and Tree of Thought: A Theoretical Framework for Sequential Reasoning and Test-Time Optimization in Transformers},
  author={Noguer i Alonso, Miquel},
  year={2024},
  howpublished={arXiv preprint arXiv:2412.XXXXX},
  note={Artificial Intelligence Finance Institute}
}

@inproceedings{sprague2025tocot,
  title={To CoT or Not To CoT? Chain-of-Thought Helps Mainly on Math and Symbolic Reasoning},
  author={Sprague, Zayne and Yin, Fangcong and Rodriguez, Juan Diego and Jiang, Dongwei and Wadhwa, Manya and Singhal, Prasann and Zhao, Xinyu and Ye, Xi and Mahowald, Kyle and Durrett, Greg},
  booktitle={Proceedings of ICLR},
  year={2025},
 
}

@inproceedings{era2024empowering,
  title={Empowering Bengali Education with AI: Solving Bengali Math Word Problems through Transformer Models},
  author={Era, Jalisha Jashim and Paul, Bidyarthi and Aothoi, Tahmid Sattar and Zim, Mirazur Rahman and Shah, Faisal Muhammad},
  booktitle={2024 27th International Conference on Computer and Information Technology (ICCIT)},
  pages={909--914},
  year={2024},
  organization={IEEE}
}

@misc{meta2024llama3,
  title        = {LLaMA 3: Open-Weight Models for Reasoning and Multilingual Tasks},
  author       = {{Meta AI}},
  year         = {2024},
  howpublished = {open-weight-models-for-reasoning-and-multilingual-tasks/}}

@article{haji2024multiagenttot,
  title   = {Improving LLM Reasoning with Multi-Agent Tree-of-Thought Validator Agent},
  author  = {Haji, Fatemeh and Bethany, Mazal and Tabar, Maryam and Chiang, Cho-Yu Jason and Rios, Anthony and Najafirad, Peyman},
  journal = {arXiv preprint arXiv:2409.11527},
  year    = {2024},
  
}

@inproceedings{touvron2023llama,
  title     = {LLaMA: Open and Efficient Foundation Language Models},
  author    = {Touvron, Hugo and Lavril, Thibaut and Izacard, Gautier and Martinet, Xavier and Lachaux, Marie-Anne and Lacroix, Timoth{\'e}e and Roziere, Baptiste and Goyal, Naman and Hambro, Eric and Azhar, Faisal and Rodriguez, Aurel}
}

@article{openai2025gptoss,
  title   = {gpt-oss-120b \& gpt-oss-20b Model Card},
  author  = {OpenAI},
  journal = {arXiv preprint arXiv:2508.10925},
  year    = {2025},
  
}

% \begin{thebibliography}{00}
% \bibitem{b1} G. Eason, B. Noble, and I. N. Sneddon, ``On certain integrals of Lipschitz-Hankel type involving products of Bessel functions,'' Phil. Trans. Roy. Soc. London, vol. A247, pp. 529--551, April 1955.
% \bibitem{b2} J. Clerk Maxwell, A Treatise on Electricity and Magnetism, 3rd ed., vol. 2. Oxford: Clarendon, 1892, pp.68--73.
% \bibitem{b3} I. S. Jacobs and C. P. Bean, ``Fine particles, thin films and exchange anisotropy,'' in Magnetism, vol. III, G. T. Rado and H. Suhl, Eds. New York: Academic, 1963, pp. 271--350.
% \bibitem{b4} K. Elissa, ``Title of paper if known,'' unpublished.
% \bibitem{b5} R. Nicole, ``Title of paper with only first word capitalized,'' J. Name Stand. Abbrev., in press.
% \bibitem{b6} Y. Yorozu, M. Hirano, K. Oka, and Y. Tagawa, ``Electron spectroscopy studies on magneto-optical media and plastic substrate interface,'' IEEE Transl. J. Magn. Japan, vol. 2, pp. 740--741, August 1987 [Digests 9th Annual Conf. Magnetics Japan, p. 301, 1982].
% \bibitem{b7} M. Young, The Technical Writer's Handbook. Mill Valley, CA: University Science, 1989.
% \end{thebibliography}

\end{document}